\def\year{2021}\relax
\documentclass[letterpaper]{article} 
\usepackage{aaai21}  
\usepackage{times}  
\usepackage{helvet} 
\usepackage{courier}  
\usepackage[hyphens]{url}  
\usepackage{graphicx} 
\usepackage{natbib}  
\usepackage{caption} 
\usepackage{adjustbox}
\usepackage{latexsym}
\usepackage{multirow}
\usepackage{diagbox}
\usepackage{amsmath}
\usepackage{pifont}
\usepackage{amssymb}
\usepackage{booktabs}
\usepackage{makecell}
\usepackage{subfigure}
\usepackage{microtype}

\newcommand{\cmark}{\ding{51}}%
\newcommand{\xmark}{\ding{55}}%

\makeatletter
\newcommand{\printfnsymbol}[1]{%
  \textsuperscript{\@fnsymbol{#1}}%
}
\makeatother

\title{AAAI Press Formatting Instructions \\for Authors Using \LaTeX{} --- A Guide }
\author{
    
    Written by AAAI Press Staff\textsuperscript{\rm 1}\thanks{With help from the AAAI Publications Committee.}\\
    AAAI Style Contributions by Pater Patel Schneider,
    Sunil Issar,  \\
    J. Scott Penberthy,
    George Ferguson,
    Hans Guesgen,
    Francisco Cruz,
    Marc Pujol-Gonzalez
    \\
}

\affiliations{
    
    $^{1}$Korea University\\
    $^{2}$Wisenut Inc.\\
    $^{3}$Electronics and Telecommunications Research Institute\\
    \tt$^{1}$\{ec\_park, limhseok\}@korea.ac.kr\\
    \tt$^{2}$taesunwhang@wisenut.co.kr, \tt$^{3}$ycyoon@etri.re.kr
    
}

\title{Multi-View Attention Network for Visual Dialog}
\author {
    Sungjin Park$^{1}\thanks{\ \ Equal Contribution} $~~~Taesun Whang$^{2}\printfnsymbol{1} \hspace{0.005em} \thanks{\ \ Work performed while at Korea University}$~~~Yeochan Yoon$^{3}$~~~Heuiseok Lim$^{1}$\\
}

\begin{document}

\maketitle

\begin{abstract}
Visual dialog is a challenging vision-language task in which a series of questions visually grounded by a given image are answered. To resolve the visual dialog task, a high-level understanding of various multimodal inputs ({\em{e.g.,}} question, dialog history, and image) is required. Specifically, it is necessary for an agent to 1) determine the semantic intent of question and 2) align question-relevant textual and visual contents among heterogeneous modality inputs. In this paper, we propose Multi-View Attention Network (MVAN), which leverages multiple views about heterogeneous inputs based on attention mechanisms. MVAN effectively captures the question-relevant information from the dialog history with two complementary modules ({\em{i.e.,}} Topic Aggregation and Context Matching), and builds multimodal representations through sequential alignment processes ({\em{i.e.,}} Modality Alignment). Experimental results on VisDial v1.0 dataset show the effectiveness of our proposed model, which outperforms the previous state-of-the-art methods with respect to all evaluation metrics.\end{abstract}

\section{Introduction}
\begin{figure}[t]\centering
\includegraphics[width=0.4\textwidth]{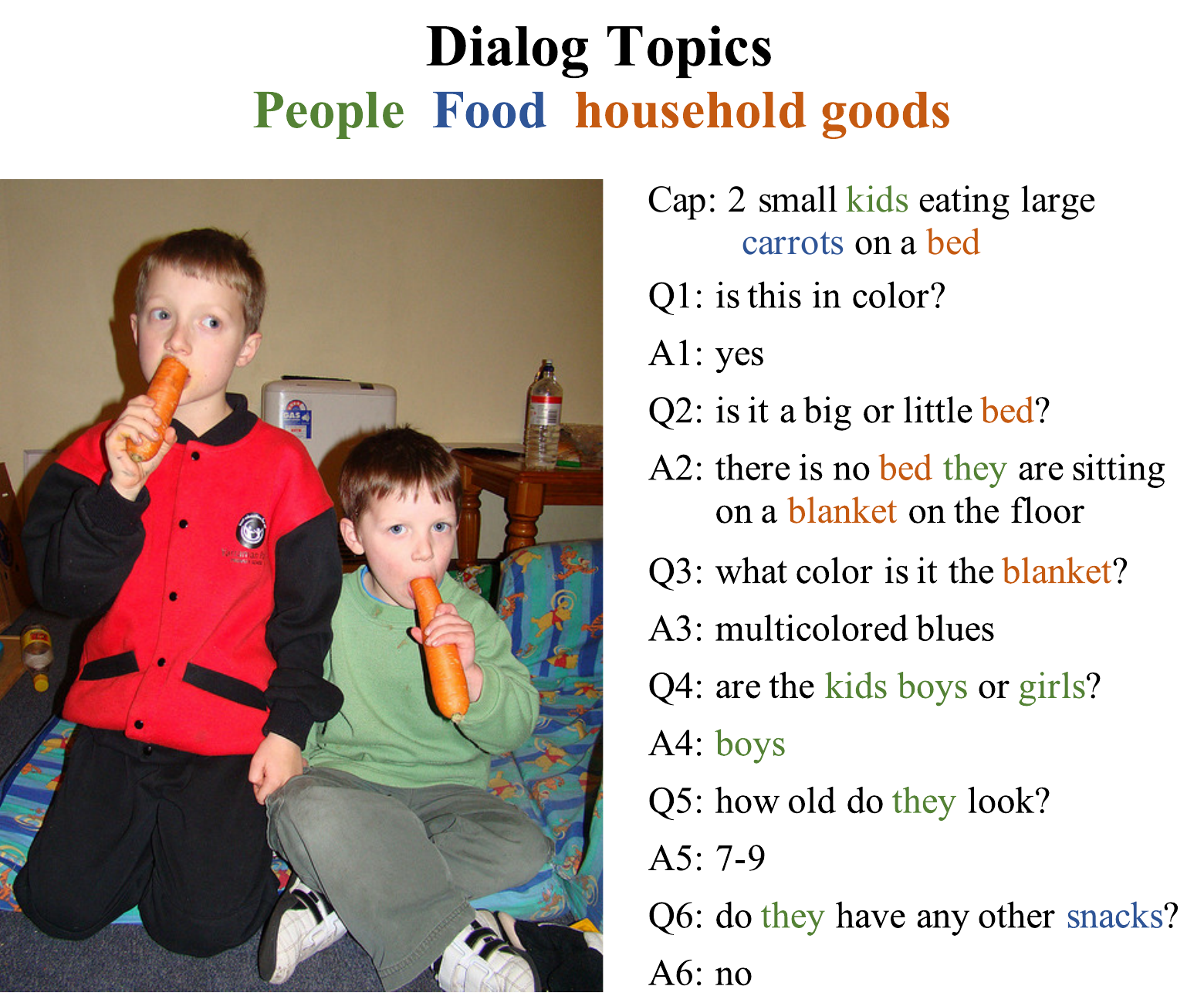}
\caption{Example of a visual dialog task. The text color indicates the dialog topic ({\em{e.g.,}} ``people'', ``food'', and ``household goods'').
}
\label{fig:fig1_visdial_example}
\end{figure}

As a part of the interdisciplinary research that combines natural language processing with computer vision, a wide variety of vision--language tasks ({\em{e.g.,}} visual question answering (VQA), image captioning, referring expressions, and etc.) have been introduced in recent years. Considerable efforts in this field have advanced the capabilities of artificial intelligence agents a step further, but the agent's comprehension of multimodal information is still far from human-level reasoning and cognitive ability \cite{hudson2019gqa,zellers2019recognition}.\\
\indent The visual dialog task is similar to VQA in that it requires the agent to answer a question that is guided by an image, but this task differs in that the agent needs to answer a series of questions focusing on previous dialog as well as a given image. It is more challenging than other vision--language tasks because the agent is asked to selectively ground the visual contents related to the question topics, which change as the dialog proceeds. For example, Figure \ref{fig:fig1_visdial_example} illustrates how the question topics change during each dialog turn ({\em{e.g.,}} ``household goods'', ``people'', and ``food''). Furthermore, answering some questions that contain ambiguous expressions ({\em{e.g.,}} question 5 (Q5): ``how old do {\em{they}} look?'') can be more difficult because the agent should consider which entity ``\textit{they}'' refers to ({\em{i.e.,}} ``kids'' and ``boys'' in caption and Q4--A4 pair) and then ground the referent in the image. To address these issues, the dialog agent should be capable of determining the semantic intent of the question, clarifying referential ambiguities, and then leveraging grounded visual contents and identified semantic information.\\
\indent Several recent researches have been performed to solve the visual dialog task from the perspective of visual co-reference resolution \cite{seo2017visual, kottur2018visual, kang2019dual, niu2019recursive}. However, resolving the visual co-reference does not always lead to complete understanding of semantic intent and topics of the question. For example, when answering Q6 in Figure \ref{fig:fig1_visdial_example}, the agent is required to not only resolve visual co-reference, but also explicitly understand the semantic intent of the question which is asking whether other ``snacks'' exist or not ({\em{i.e.,}} the model should focus on ``snacks'' rather than ``\textit{they}''). From these observations, it is crucial to capture what the topic of the question is in order to accurately determine the semantic intent of the question. \\
\indent To this end, this paper proposes Multi-View Attention Network (MVAN), which leverages question-guided contextual information and clues through the dialog history; and then effectively learns semantic alignments between visual and textual representations through the sequential alignment processes. MVAN consists of three main modules. First, the Context Matching module effectively represents contextual information of dialog history that is relevant to the question at sentence level. This is because, in general, the semantic intent of a sentence tends to be determined by the context of the entire sequence as well as some words that are directly connected to the topic. Second, the Topic Aggregation module is capable of capturing topic-guided clues from the dialog history at word level. This takes advantage of the fact that topic-related representation is well-constructed by directly using the original word embeddings of each word. Both modules adaptively propagate textual information that interacts with the semantic intent of the current question by attention mechanism and gate function. Lastly, the Modality Alignment module performs two sequential aligning processes to learn semantic alignments between each textual output of previous modules with visual features. Since the alignments between contextual representation and visual contents can be implicit and noisy, the Modality Alignment module first learns the soft mapping between heterogeneous inputs based on the topic-guided clues, then sequentially aligns them with the surrounding contextual information. \\
\indent The main contributions of this paper are as follows. 1) We propose MVAN that consists of two complementary views and combine them with visual contents through multiple alignment steps. 2) Experimental results on VisDial v1.0 show that our proposed model outperforms the previous state-of-the-art methods with respect to all evaluation metrics. 3) Visualization of reasoning process for each module demonstrates that MVAN explicitly understands the semantic intent of the question, which leads to a reasonable interpretation of employing various multimodal inputs.

\section{Related Work}
\begin{figure*}[t]\centering
\includegraphics[width=0.86\textwidth]{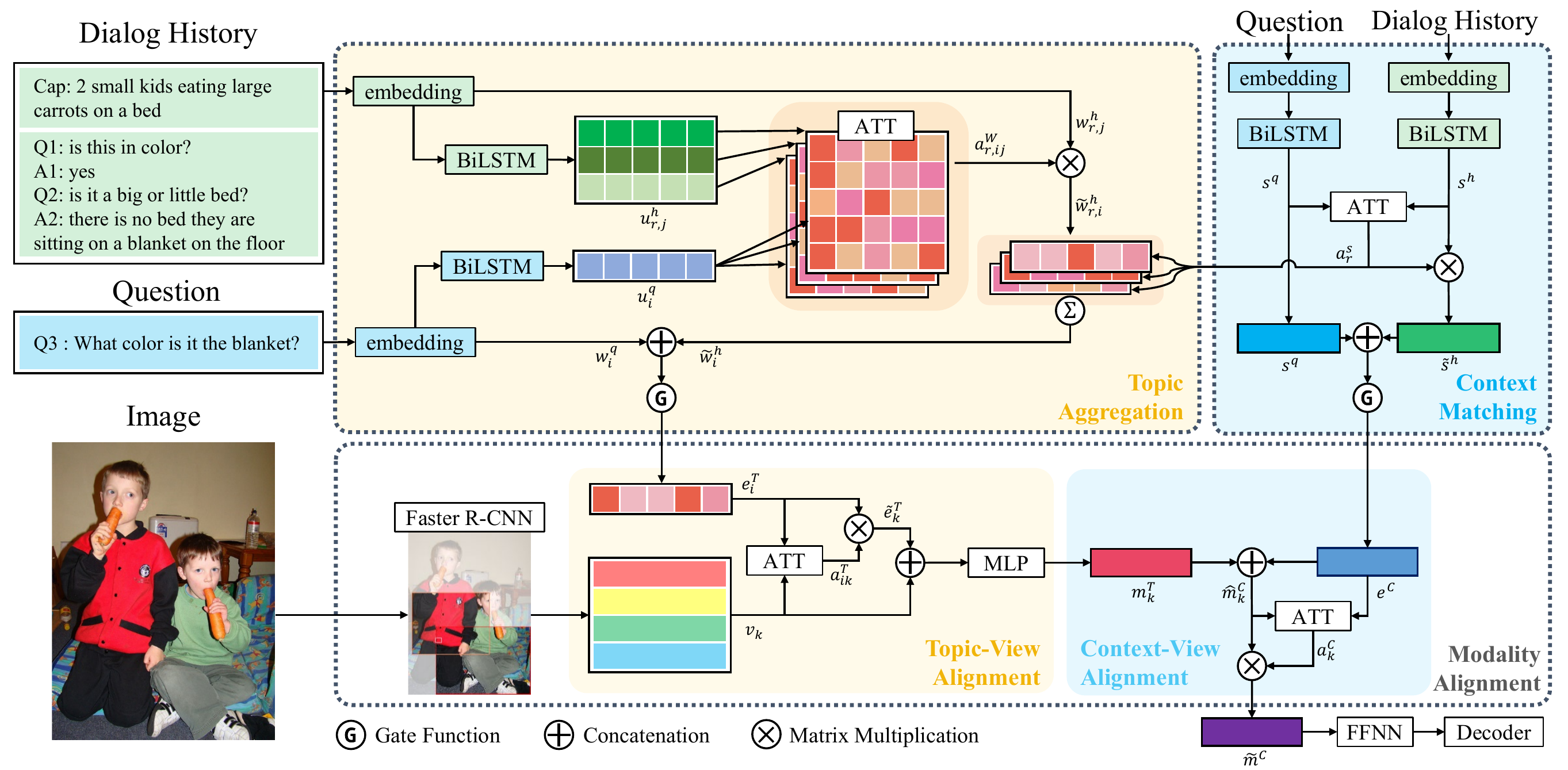}
\caption{Model architecture of Multi-View Attention Network (MVAN). BiLSTM layers for the question (blue) and the dialog history (green) are shared in Topic Aggregation Module and Context Matching Module, respectively.}
\label{fig:fig2_mvan}
\end{figure*}

Visual dialog is a task proposed by \citet{das2017visual} that requires the dialog agent to answer the current question by exploiting both the image and dialog history. \citet{das2017visual} also introduced encoder-decoder models such as late fusion, hierarchical recurrent network, and memory network as baseline methods. Most of the previous approaches are predominantly based on attention mechanisms to fuse representations of the given multimodal inputs \cite{wu2018you, guo2019image, gan2019multi, kim2020modality}. Another direction of research that is inspired by graphical networks focuses on learning the inherent relations among image, dialog history, and question \cite{zheng2019reasoning, schwartz2019factor, guo2020iterative}.\\
\indent On the other hand, several approaches that explicitly resolve ambiguous references are based on visual co-reference resolution. \citet{kottur2018visual} used neural module networks \cite{andreas2016neural} to effectively link references and ground relevant visual contents  at word level. \citet{kang2019dual} adapted a self-attention mechanism \cite{vaswani2017attention} based on sentence-level representation to resolve referential ambiguity. \citet{niu2019recursive} proposed a recursive attention mechanism to capture question-relevant dialog history and ground related visual contents to the image. 
Most existing works \cite{kottur2018visual,kang2019dual} that use only word or sentence representations have limitations in identifying the semantic intent of the question. Unlike this, our proposed model considers both topic-related clues and contextual information to effectively capture the semantic intent of the question. In addition, MVAN adaptively integrates dialog history and visual contents by performing sequential alignment steps rather than exploiting only dialog history and visual contents that meet specific recursion conditions \cite{niu2019recursive}. \\
\indent More recently, \citet{murahari2019large} and \citet{wang2020vd} introduced a fine-tuning method using a pre-trained model, as it is observed that pre-trained language model architectures ({\em{e.g.,}} BERT \cite{devlin2019bert}) also effectively perform on vision-language tasks \cite{lu2019vilbert}. Also, \citet{qi2019two} proposed causal intervention algorithms that can be applicable to other visual dialog models and \citet{agarwal2020history} proposed curriculum fine-tuning inspired by the work of \citet{bengio2009curriculum}.

\section{Model}
\label{sec:model}
In the visual dialog task \cite{das2017visual}, a dialog agent is given a set of multimodal inputs for each dialog turn $t$. This input set consists of an image $I$, a current question $Q_{t}$, the dialog history set $H_{t}=\{C,(Q_1,A_1^{gt}),\cdots,(Q_{t-1},A_{t-1}^{gt})\}$, which contains image caption $C$ and $t-1$ consecutive question-answer pairs, and a set of answer candidates $A_{t} = \{A_{t}^{1}, A_{t}^{2}, ..., A_{t}^{100}\}$. The agent then is required to answer the question by either discriminating or generating a correct answer.
\subsection{Multimodal Representation}
\paragraph{Visual Features} \indent We employ a bottom-up attention mechanism \cite{anderson2018bottom} to represent the objects appearing in an image. Visual features of object regions $\{v_k\}_{k=1}^{n_v} \in \mathbb{R}^{d_{v} \times n_v}$, where $n_v$ is the number of detected objects ranging from 10 to 100, are adaptively extracted from a Faster-RCNN \cite{ren2015faster} that is pre-trained with Visual Genome \cite{krishna2017visual}.
\paragraph{Language Features} \indent We first embed three different text inputs: the current question, dialog history, and answer candidates. The word embedding layer is initialized with pre-trained GloVe embeddings \cite{pennington2014glove}. We then feed the word embeddings into a bi-directional long short-term memory (BiLSTM) to encode a sequential representation of each embedding. Specifically, each word in question $Q$ is embedded as $\{w_{i}^{q}\}_{i=1}^{n_q}\in \mathbb{R}^{d_{w} \times n_q}$, where $n_q$ is the number of words in the sequence. Each word embedding is fed into the BiLSTM layer as follows:
\begin{subequations}\label{eq:bilstm}
\begin{align}
\overrightarrow u_{i}^q &=\textnormal{LSTM}(w_{i}^{q}, \overrightarrow u_{i-1}^q)\\
\overleftarrow u_{i}^q &= \textnormal{LSTM}(w_{i}^{q}, \overleftarrow u_{i+1}^q).
\end{align}
\end{subequations}
The sequential representation of each token is constructed by concatenating the hidden states of the forward and backward LSTMs, denoted as $u_{i}^q=[\overrightarrow u_{i}^q , \overleftarrow u_{i}^q]$. Meanwhile, the sequential representation for each dialog history $u^{h}_{r}=\{u_{r,j}^{h}\}_{j=1}^{n_{h}} (0\leq r \leq t-1)$ is constructed using the question construction process with different BiLSTM layers. For the answer candidates, we use a different uni-directional LSTM to represent them because their sequence lengths are shorter than those of the questions.

\subsection{Multi-View Attention Network}
We propose MVAN that considers the semantic intent and topic of the question simultaneously and effectively aligns the textual and visual information through multiple alignment processes. Figure \ref{fig:fig2_mvan} describes the MVAN architecture, which consists of three components: 1) Context Matching, 2) Topic Aggregation, and 3) Modality Alignment.

\paragraph{Context Matching Module} \indent Generally, the semantic intent of a sentence not only relies on particular words that implicitly point to the topic of the sentence but also tends to be determined by the context of the entire sequence. Therefore, we build the Context Matching module that adaptively integrates the question and its relevant history at sentence level. Contextual representation is constructed by concatenating the last hidden states of the forward and backward LSTMs for the question and dialog history, denoted as $s^{q} = [\overrightarrow u_{1}^q , \overleftarrow u_{n_{q}}^{q}]$ and $s^{h}_{r} = [\overrightarrow u_{r,1}^{h} , \overleftarrow u_{r,n_{h_r}}^{h}]$, respectively. We then apply an attention mechanism to focus on question-relevant history. The Context Matching module takes contextual representation of the question $s^q\in \mathbb{R}^{d_{s}\times1}$ and dialog history $s^{h}=\{s^h_0,s^h_1,...,s^h_{t-1}\}\in \mathbb{R}^{d_{s} \times t}$ and outputs question-relevant history features as follows:
\begin{subequations}\label{eq:context_matching}
\begin{align}
z_{r}^\mathcal{S} &= \mathbf{W}^\top(f_{q}^\mathcal{S}(s^q)\circ f_{h}^\mathcal{S}(s^{h}_{r})) + b\\
a_{r}^\mathcal{S} &= \text{softmax}(z_{r}^\mathcal{S}) \\
\tilde s^{h} &= \textstyle \sum\nolimits_{r=0}^{t-1}a_{r}^\mathcal{S}s^{h}_{r},
\end{align}
\end{subequations}
where $\circ$ is element-wise multiplication, $\mathbf{W}\in \mathbb{R}^{d_f \times 1}$ is a projection matrix, and $f_{q}^\mathcal{S}(\cdot)$ and $f_{h}^\mathcal{S}(\cdot)$ denote non-linear transformation functions. We apply a gate function to automatically filter out  dialog history that is irrelevant to current question as follows:
\begin{subequations}\label{eq:context_matching_gate}
\begin{align}
gate^{\mathcal{C}}&=\sigma ({\mathbf{W}^\mathcal{C}_{gate}}[s^q,\tilde s^{h}] + b^{\mathcal{C}}_{gate})\\
e^{\mathcal{C}}&= gate^{\mathcal{C}} \circ [s^q,\tilde s^{h}],
\end{align}
\end{subequations}
where $\sigma(\cdot)$ is a sigmoid function and $\mathbf{W}^\mathcal{C}_{gate}\in \mathbb{R}^{2d_{s} \times 2d_{s}}$ and $b^\mathcal{C}_{gate}\in \mathbb{R}^{2d_{s} \times  1}$ are trainable parameters. Note that $e^\mathcal{C}\in\mathbb{R}^{2d_{s} \times 1}$ is a context-matching representation that selectively combines the contextual information of the question and question-relevant dialog history.
\paragraph{Topic Aggregation Module} \indent The topic of the question is generally expressed in a single word or phrase and likely to be connected with clues ({\em{i.e.,}} the topic of previous questions in the dialog history). We design the Topic Aggregation module to combine the clues associated with the question topic by exploiting the initial word embeddings ({\em{i.e.,}} GloVe) to represent their original meaning. Specifically, this module leverages word-level sequential representation of the question and dialog history, $\{u^q_i\}_{i=1}^{n_q}\in\mathbb{R}^{d_u \times n_q}$ and $\{u_{r,j}^{h}\}_{j=1}^{n_h}\in\mathbb{R}^{d_u \times n_h}$, respectively. The dot product attention mechanism is employed to selectively focus the words that are relevant to the question topic from the dialog history as follows:
\begin{subequations}\label{eq:topic_aggregation}
\begin{align}
z_{r,ij}^\mathcal{W} &= f_{q}^{\mathcal{W}}(u^q_i)^\top f_{h}^\mathcal{W}(u^{h}_{r,j})\\
a_{r,ij}^\mathcal{W} &= exp(z_{r,ij}^\mathcal{W})/ \textstyle \sum\nolimits_{j=1}^{n_h}exp(z_{r,ij}^\mathcal{W}) \\
\tilde w_{r,i}^{h} &= \textstyle \sum\nolimits_{j=1}^{n_r}a_{r,ij}^\mathcal{W}w_{r,j}^h,
\end{align}
\end{subequations}
where $f_{q}^\mathcal{W}(\cdot)$ and $f_{h}^\mathcal{W}(\cdot)$ are non-linear transformation functions. The question-guided history feature for each round $\tilde w_{r,i}^{h}$ is computed by a weighted sum of their word embeddings, which represent the original meanings of words. The attended representation $\{\tilde w_{i}^{h}\}_{i=1}^{n_q}\in\mathbb{R}^{d_w \times n_q}$ is computed by aggregating overall all history $\{\tilde w_{r,i}^{h}\}_{r=0}^{t-1}$, weighted by the attention scores of the Context Matching module $a_r^\mathcal{S}$ as follows: 
\begin{equation}
\tilde w_{i}^{h} = \textstyle \sum\nolimits_{r=0}^{t-1}a_{r}^\mathcal{S}\tilde w_{r,i}^{h}.
\label{eq:word_view_weighted_sum}
\end{equation} 
Similar to the Context Matching module, the gate operation adaptively filters out irrelevant topic-guided clues at word level.
\begin{subequations}\label{eq:topic_aggregation_gate}
\begin{align}
gate^{\mathcal{T}}_i&=\sigma ({\mathbf{W}^\mathcal{T}_{gate}}[w^q_i,\tilde w^h_i] + b^{\mathcal{T}}_{gate})\\
e^{\mathcal{T}}_i&= gate^{\mathcal{T}}_i \circ [w^q_i,\tilde w^h_i],
\end{align}
\end{subequations}
where $\mathbf{W}^{\mathcal{T}}_{gate}\in \mathbb{R}^{2d_{w} \times 2d_{w}}$ and $b^{\mathcal{T}}_{gate}\in\mathbb{R}^{2d_{w} \times 1}$ are trainable parameters. Note that $\{e_i^{\mathcal{T}}\}_{i=1}^{n_q}$ $\in\mathbb{R}^{2d_w\times n_q}$ is topic-aggregation representations that encode adaptive information of the question topic and various history clues associated with it.

\paragraph{Modality Alignment Module} \indent Given the output representations of the Context Matching module $e^{\mathcal{C}}\in\mathbb{R}^{2d_s \times 1}$ and Topic Aggregation module $\{e_i^{\mathcal{T}}\}_{i=1}^{n_q}\in\mathbb{R}^{2d_w\times n_q}$, the Modality Alignment module aligns them with visual features $\{v_k\}_{k=1}^{n_v} \in \mathbb{R}^{d_{v} \times n_v}$ via two-step alignments. First, topic-view alignment performs soft alignment between heterogeneous modalities at topic level before mapping high-level contextual representation with the visual features. We utilize dot-product attention to represent the visual relevant topic-aggregation embeddings as follows:
\begin{subequations}\label{eq:modality_fusion}
\begin{align}
z_{ik}^\mathcal{T} &= f_{\ell}^{\mathcal{W}}(e^\mathcal{T}_i)^\top f_{v}^\mathcal{T}(v_k)\\
a_{ik}^\mathcal{T} &= exp(z_{ik}^\mathcal{T})/ \textstyle \sum\nolimits_{i=1}^{n_q}exp(z_{ik}^\mathcal{T}) \\
\tilde e_{k}^{\mathcal{T}} &= \textstyle \sum\nolimits_{i=1}^{n_q}a_{ik}^\mathcal{T}e_{i}^\mathcal{T},
\end{align}
\end{subequations}
where $f_{\ell}^{\mathcal{W}}(\cdot)$ and $f_{v}^\mathcal{W}(\cdot)$ are non-linear transformation functions to embed two different modality representations into the same embedding space. Then, we obtain the fused feature vectors by concatenating the visual and attended topic-aggregation features and using a multi-layer perceptron (MLP).
\begin{equation}
m_{k}^{\mathcal{T}} = \text{MLP}([v_{k},\tilde{e}_{k}^{\mathcal{T}}])
\label{eq:mlp}
\end{equation} 
Note that $\{m_{k}^{\mathcal{T}}\}_{k=1}^{n_v}\in\mathbb{R}^{d_v\times n_v}$ is a topic-view aligned representation that is aligned information across the visual contents with the salient question topics.
This is passed to the context-view aligning step as follows:
\begin{subequations}
\label{eq:context_view_fusion}
\begin{align}
\hat m_{k}^{\mathcal{C}}&=[m_{k}^{\mathcal{T}},e^\mathcal{C}]\\
z_k^\mathcal{C} &= \text{L2Norm}(f_{m}^{\mathcal{C}}(\hat m_{k}^{\mathcal{C}}) \circ f_{\ell}^{\mathcal{C}}(e^{\mathcal{C}}))\\
a_k^\mathcal{C}&=\text{softmax}(\mathbf{W}^{\top}z_k^{\mathcal{C}}+b) \\
\tilde m^{\mathcal{C}} &= \textstyle \sum\nolimits_{k=1}^{n_v}a_{k}^\mathcal{C} \hat m_{k}^{\mathcal{C}},
\end{align}
\end{subequations}
\noindent where $f_{m}^{\mathcal{C}}(\cdot)$ and $f_{\ell}^\mathcal{C}(\cdot)$ are non-linear transformation functions and $\mathbf{W}\in\mathbb{R}^{d_f\times 1}$ is a projection matrix. Note that $\tilde m^{\mathcal{C}}\in\mathbb{R}^{(d_v + 2d_s) \times 1}$ is a context-view aligned representation, which is realigned using context-matching representation. These multiple alignment processes allow the model to understand the semantic intent of the question with complementary views, and effectively align the corresponding heterogeneous multimodal inputs. Finally, this enhanced feature is fed into a single-layer feed-forward neural network (FFNN) with a ReLU activation function. 

\begin{equation}
m^{enc}=\text{max}(0, \mathbf{W}^\top\tilde m^{\mathcal{C}} + b),
\label{eq:multimodal_representation}
\end{equation}
where $\mathbf{W}\in\mathbb{R}^{(d_v+2d_s) \times d_{enc}}$ and $b\in\mathbb{R}^{d_{enc} \times 1}$ are trainable parameters. Note that $m^{enc}\in\mathbb{R}^{d_{enc}\times 1}$ is the multi-view aligned representation, which is fed into either a discriminative or generative decoder.

\subsection{Answer Decoder}
\label{sec:answer_decoder}
\paragraph{Discriminative Decoder} \indent We use the last hidden states of the forward LSTM to encode sentence representations of answer candidates, denoted as $s^{a}=\{\overrightarrow u^a_{i,n_a}\}_{i=1}^{100}\in\mathbb{R}^{d_a\times 100}$. We rank them according to the dot products of the candidates $s^a$ and multi-view aligned representation $m^{enc}$, then apply the softmax function to obtain the probability distribution of the candidates, denoted as $p = \text{softmax}((s^{a})^\top m^{enc})$. 
Note that dimension of each answer candidate representation is same as that of the encoder output. We use multi-class cross entropy loss as the discriminative objective function, formulated as
\begin{equation}
\mathcal{L}_{D}=-\textstyle\sum\nolimits_{i=1}^{100}y_{i}\text{log}p_i,
\label{eq:discriminative_loss}
\end{equation}
where $y_i$ is a one-hot encoded vector of the ground truth answer.

\paragraph{Generative Decoder} \indent Unlike most previous approaches, which take only a discriminative approach, we also train our model in a generative manner \cite{das2017visual}. During the training phase, we use a two-layer LSTM to predict the next token given the previous tokens in the answer sequence. The initial hidden state of the LSTM is initialized with the encoder output representation. For each answer candidate, we compute the likelihood of the ground truth of each token, denoted as $\{p_k\}_{k=1}^{n_a}$, and train the model by minimizing the summation of the negative log-likelihood as follows:
\begin{equation}
\mathcal{L}_{G}=-\textstyle\sum\nolimits_{k=1}^{n_a}\text{log}p_k.
\label{eq:generative_loss}
\end{equation}

\begin{table*}[ht]\centering
\begin{adjustbox}{width=0.9\textwidth}

\begin{tabular}{l|cccccc|ccccc}   
\toprule                     
\multirow{2}{*}{Model} &\multicolumn{6}{c|}{VisDial v1.0 (test-std)} & \multicolumn{5}{c}{VisDial v0.9 (val)}  \\ 
& NDCG & MRR & R@1 & R@5 & R@10 & Mean & MRR & R@1 & R@5 & R@10 & Mean \\
\midrule
LF \cite{das2017visual} &45.31 &55.42 &40.95 &72.45 &82.83 &5.95 &58.07 &43.82 &74.68 &84.07 &5.78  \\ 
HRE \cite{das2017visual} &45.46 &54.16 &39.93 &70.45 &81.50 &6.41 &58.46 &44.67 &74.50 &84.22 &5.72\\ 
MN \cite{das2017visual} &47.50 &55.49 &40.98 &72.30 &83.30 &5.92 &59.65 &45.55 &76.22 &85.37 &5.46 \\ 
HCIAE \cite{lu2017best} &- &- &- &- &- &- &62.22 &48.48 &78.75 &87.59 &4.81\\ 
AMEM \cite{seo2017visual} &- &- &- &- &- &- &62.27 &48.53 &78.66 &87.43 &4.86\\ 
CoAtt \cite{wu2018you} &- &- &- &- &- &- &63.98 &50.29 &80.71 &88.81 &4.47\\ 
FGA \cite{schwartz2019factor} &52.10 &63.70 &49.58 &\underline{80.98} &88.55 &4.51 &67.12 & 54.02 & 83.21 & 90.47 & 4.08 \\
CorefNMN \cite{kottur2018visual} &54.70 &61.50 &47.55 &78.10 &88.80 &4.40 &64.10 &50.92 &80.18 &88.81 &4.45\\ 
RVA \cite{niu2019recursive} & 55.59 & 63.03 & 49.03 & 80.40 & 89.83 & 4.18 & 66.34 & 52.71 & 82.97 & 90.73 & 3.93 \\
DualVD \cite{jiang2020dualvd} &56.32 &63.23 &49.25 &80.23 &89.70 &\underline{4.11} &62.94 &48.64 &80.89 &89.94 &4.17 \\ 
CAG \cite{guo2020iterative} &56.64 &63.49 &49.85 &80.63 &\underline{90.15} &\underline{4.11} &67.56 &54.64 & \underline{83.72} & \textbf{\underline{91.48}} & \underline{3.75} \\ 
HACAN \cite{yang2019hacan} &57.17 & \underline{64.22} & \underline{50.88} &80.63 &89.45 &4.20 &  \textbf{\underline{67.92}} & \textbf{\underline{54.76}} & 83.03 & 90.68 & 3.97 \\ 
Synergistic \cite{guo2019image} & 57.32 & 62.20 & 47.90 & 80.43 & 89.95 & 4.17 & - & - & - & - & - \\ 
DAN \cite{kang2019dual} & \underline{57.59} & 63.20 & 49.63 & 79.75 & 89.35 & 4.30 & 66.38 & 53.33 & 82.42 & 90.38 & 4.04 \\ 
\midrule
\textbf{MVAN}  &\textbf{59.37} &\textbf{64.84} &\textbf{51.45} &\textbf{81.12} &\textbf{90.65} &\textbf{3.97} &67.65 &54.65 &\textbf{83.85} & 91.47 & \textbf{3.73} \\
\textbf{MVAN$^{\dagger}$} &60.92 &66.38 &53.20 &82.45 &91.85 &3.68 & 69.35 & 56.59 & 85.29 & 92.53 & 3.43 \\ 
\bottomrule

\end{tabular}
\end{adjustbox}
\caption{Results on VisDial v1.0 (test-std) and v0.9 (val). $\dagger$ denotes ensembles.}
\label{tab:results_visdial}
\end{table*}

\paragraph{Multi-task Learning} \indent 
We perform multi-task learning by combining both discriminative and generative decoders to classify answers, denoted as $\mathcal{L}=\mathcal{L}_{D}+\mathcal{L}_{G}$.
For the evaluation, we simply average the probability distributions of each decoder. Multi-task learning substantially improves performance with respect to the normalized discounted cumulative gain (NDCG) metric.

\section{Experiments}
\subsection{Experimental Setup}
\paragraph{Datasets} \indent We use the VisDial v0.9 and v1.0 datasets to evaluate our proposed model. VisDial v0.9 \cite{das2017visual} consists of 123k MS-COCO \cite{lin2014microsoft} images and their captions. The training and validation splits of VisDial v0.9 contain 83k and 40k images respectively, and each image has 10 consecutive question-answer pairs. VisDial v1.0, which was released by supplementing VisDial v0.9, has 123k images for training splits that combine the training and validation splits of VisDial v0.9. An additional 10k images from the Flickr dataset are utilized to construct the validation and test splits in VisDial v1.0, which contain 2k and 8k images, respectively. Unlike the previous version of the dataset, dense annotations for each candidate answer are added in the validation and test splits.

\paragraph{Evaluation Metrics} \indent We evaluated our proposed model using several retrieval metrics, following the work of \citet{das2017visual}: 1) mean rank of the ground truth response (Mean), 2) recall at $k$ ($k$=\{1,5,10\}), which is denoted as R@$k$ and evaluates where the ground truth is positioned in the sorted list, and 3) mean reciprocal rank (MRR) \cite{voorhees1999trec}. NDCG was also introduced as a primary metric in the VisDial v1.0 dataset, and decreases when the model gives a low ranking to candidate answers with high relevance scores. MRR evaluates the precision of the model by ranking where a ground truth answer is positioned, whereas NDCG evaluates relative relevance of the predicted answers.

\paragraph{Training Details} \indent Our model is implemented using PyTorch framework \cite{paszke2019pytorch} based on open source code\footnote{\begin{footnotesize}\url{https://github.com/batra-mlp-lab/visdial-challenge-starter-pytorch}\end{footnotesize}} from the work of \citet{das2017visual}. The question and dialog history are represented using different BiLSTMs with 512 hidden states. The maximum sequence lengths of the question and dialog history are set to 20 and 40, respectively. We set the batch size to 32 and apply the Adam optimizer \cite{kingma2014adam} with an initial learning rate of 1e-5, which is gradually increased to 1e-3 until epoch 2, then decay at epochs 6 and 7 with the decay rate 0.1. Our code is publicly available\footnote{\begin{footnotesize}\url{https://github.com/taesunwhang/MVAN-VisDial}\end{footnotesize}}. 

\subsection{Quantitative Results}
\paragraph{Baselines} \indent We compare the results of our proposed model with previously published results on the VisDial v1.0 and v0.9 datasets for the following methods: LF \cite{das2017visual}, HRE \cite{das2017visual}, MN \cite{das2017visual}, HCIAE \cite{lu2017best}, AMEM \cite{seo2017visual}, CoAtt \cite{wu2018you}, FGA \cite{schwartz2019factor}, CorefNMN \cite{kottur2018visual}, RVA \cite{niu2019recursive}, DualVD \cite{jiang2020dualvd}, CAG \cite{guo2020iterative}, HACAN \cite{yang2019hacan}, Synergistic \cite{guo2019image}, and DAN \cite{kang2019dual}.

\begin{table}[t]\centering
\begin{adjustbox}{width=0.45\textwidth}

\begin{tabular}{l|cccccc} 
\toprule       
Model & NDCG & MRR & R@1 & R@5 & R@10 & Mean \\
\midrule
ReDAN &61.86 &53.13 &41.38 &66.07 &74.50 &8.91 \\
LTMI$^{\ddagger}$ &60.92 &60.65 &47.00 &77.03 &87.75 &4.90 \\
\midrule
MVAN$^{\ddagger}$ &\textbf{63.15} &\textbf{63.02} &\textbf{49.43} &\textbf{79.48} &\textbf{89.40} &\textbf{4.38} \\
\bottomrule
\end{tabular}
\end{adjustbox}
\caption{Results of different methods of combining discriminative and generative models on VisDial v1.0 (test-std). $\ddagger$ indicates that the model was trained using multi-task learning.}
\label{tab:results_multitask}
\end{table}


\begin{table*}[t]\centering
\begin{adjustbox}{width=0.7\textwidth}

\begin{tabular}{l|cc|cccccc}   
\toprule                     
Model 
& \multicolumn{1}{c}{\begin{tabular}[c]{@{}c@{}} context-level \\ history \end{tabular}} 
& \multicolumn{1}{c|}{\begin{tabular}[c]{@{}c@{}} topic-level \\ history\end{tabular}} 
& NDCG & MRR & R@1 & R@5 & R@10 & Mean \\
\midrule
\multirow{2}{*}{MVAN}  & \cmark & \cmark &60.17 &\textbf{65.33}	&\textbf{51.86}	&\textbf{82.40}	&\textbf{90.90}	&\textbf{3.88} \\
& \xmark  & \xmark  &\textbf{62.33}	&61.79	&47.61	&79.30	&88.81	&4.42 \\
\midrule
\multirow{2}{*}{w/o Topic Aggregation} &\cmark & N/A &58.50	&64.63	&50.84	&81.64	&90.50	&3.97  \\
& \xmark & N/A &60.57	&61.32	&47.19	&78.59	&88.40	&4.55 \\
\midrule
\multirow{2}{*}{w/o Context Matching} & N/A & \cmark &57.06	&64.15	&50.51	&81.15	&89.83	&4.12 \\
& N/A & \xmark  &58.60	&60.36	&46.09	&77.71	&87.64	&4.73\\
\bottomrule

\end{tabular}  
\end{adjustbox}
\caption{Ablations of our approaches on the VisDial v1.0 validation dataset. 
}
\label{tab:visdial1.0_ablation}
\end{table*}

\paragraph{Results on VisDial v1.0 and v0.9} \indent Table \ref{tab:results_visdial} reports the quantitative results on the VisDial v1.0 and v0.9 under the discriminative decoder setting. For VisDial v1.0, our MVAN model outperforms the previous state-of-the-art methods with respect to all evaluation metrics. 
Specifically, MVAN achieves significant improvements in NDCG from 57.59 to 59.37 and in MRR from 64.22 to 64.84, compared to the state-of-the-art baseline. In addition, we obtain better results for Mean from 4.11 to 3.97 and for R@k increased by approximately 0.4\%. Similar results are obtained in the R@5 and Mean for VisDial v0.9. We also report the results for an ensemble of 10 independent models that were trained with random initial seeds, which yields average performance improvements of 1.3\% for all metrics.\\ 
\indent These results indicate that MVAN not only has accurate prediction ability, as indicated by the non-NDCG metric results ({\em{i.e.,}} MRR, R@k, and Mean), but it has a powerful generalization capability given the result of NDCG score because this metric considers several relevant answers to be correct.

\begin{figure}[t]\centering
\includegraphics[width=0.45\textwidth]{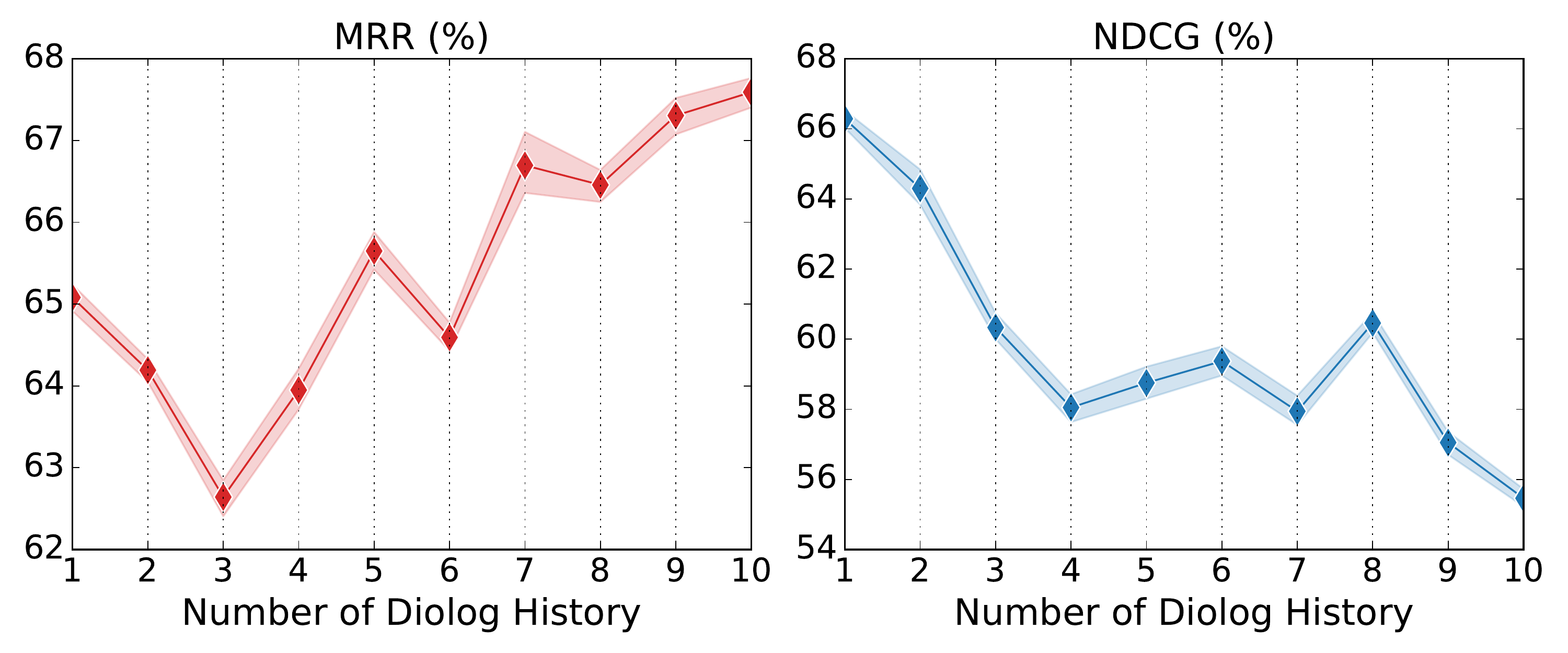}
\caption{Performance of MVAN with different amounts of dialog history on the VisDial v1.0 validation set.}
\label{fig:fig3_quantity_analysis}
\end{figure}

\paragraph{Results on multi-task learning} \indent As shown in Table \ref{tab:results_multitask}, we report the results of our MVAN model, which was trained using multi-task learning. Our proposed approach performs better with respect to all metrics than ReDAN \cite{gan2019multi}, which averages the ranking results of the discriminative and generative model, and LTMI \cite{nguyen2019efficient}, which employs multi-task learning but uses only discriminative decoder outputs for evaluation.

\paragraph{Number of dialog history}
\indent We experimented with the amount of dialog history to evaluate the impact of dialog history on the model performance in the two major metrics ({\em{i.e.,}} MRR and NDCG). The results in Figure \ref{fig:fig3_quantity_analysis} show that as the amount of dialog history information increases, the MRR tends to gradually improve, but the NDCG score deteriorates.
This result of the quantity analysis show that history information decreases the NDCG score but substantially boosts the other metrics.

\subsection{Ablation study}
We conducted ablation studies on the VisDial v1.0 validation splits to evaluate the influence of each component in our model. Modality Alignment module is not ablated because this module handles the visual features. We use the same discriminative decoder model for all ablations to exclude the impact of multitask learning.

\indent In Table \ref{tab:visdial1.0_ablation}, the first rows of each block indicate the impact of each module in our model. Because the two modules ({\em{i.e.,}} Context Matching and Topic Aggregation) are interdependent, we employ simple visual features instead of topic-aggregation representation for \textbf{MVAN w/o Topic Aggregation}, whereas we simply remove context-matching representation for \textbf{MVAN w/o Context Matching}. Both models obtain slightly lower performance with respect to all evaluation metrics than \textbf{MVAN}. We can hence infer that the two modules are complementary with respect to each other and our model integrates these complementary characteristics well for the task.\\
\indent Recent approaches \cite{murahari2019large,kim2020modality,nguyen2019efficient} reported that they observed a trade-off relationship between two primary metrics ({\em{i.e.,}} NDCG and MRR) in the visual dialog task. We also found the trade-off relationship through ablative experiments with and without dialog history features (see Table \ref{tab:visdial1.0_ablation}).  Specifically, adding dialog history features improves the MRR score by 3.54\% on average, whereas NDCG score is decreased by 1.92\% on average. 
We observe that the model has a tendency to predict the answers more precisely ({\em{i.e.,}} it has a better MRR score) when the dialog history features are added. This may imply that question-related clues in the dialog history are important factors in reasoning the ground truth, but they hinder the model's generalization ability ({\em{i.e.,}} they lower the NDCG score).

\begin{figure*}[t]\centering
\includegraphics[width=0.93\textwidth]{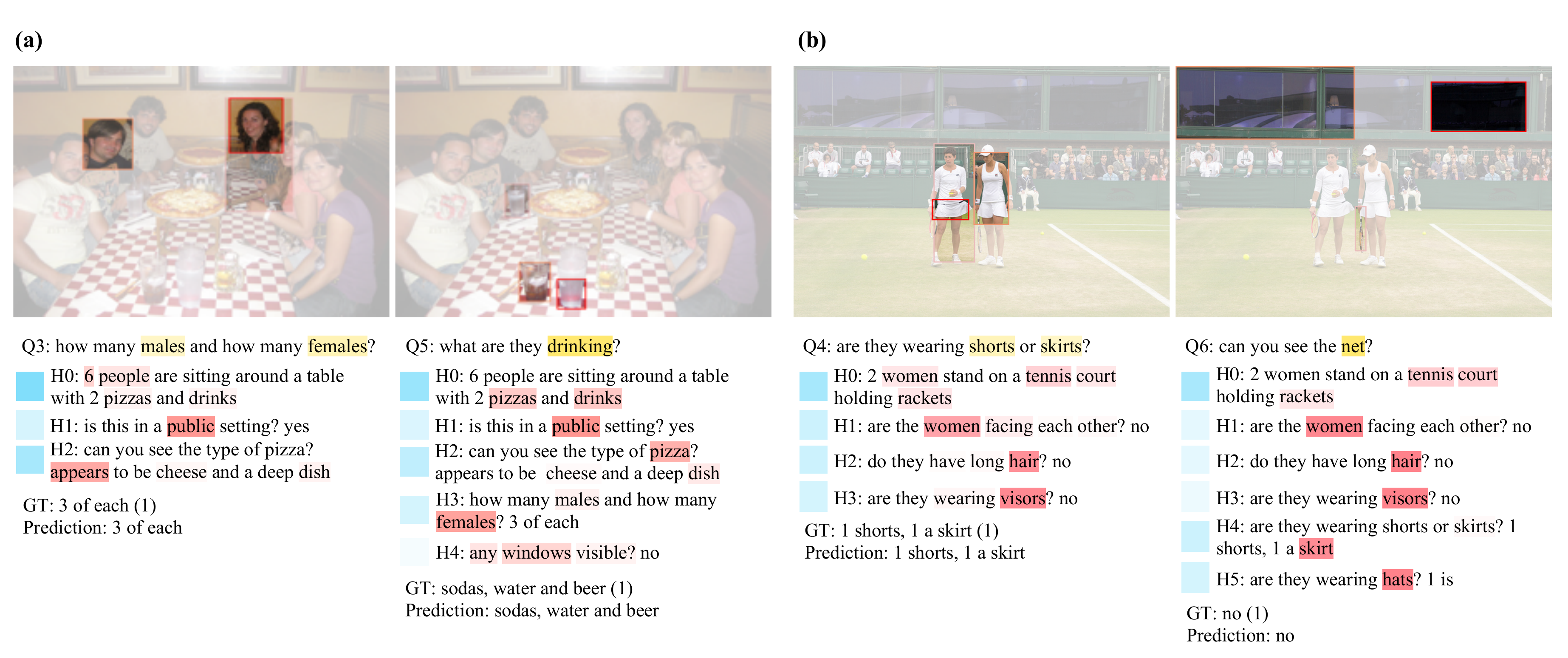}
\caption{Qualitative results on the VisDial v1.0 validation set. We visualize the different attention scores for each module: 1) attention scores from Topic Aggregation module and Context Matching module are highlighted in red and blue, respectively; 2) semantic intent of the current question represented via the topic-view alignment step in yellow; and 3) the top three attention scores of visual features from the context-view alignment step, which are represented by the b-boxes with fine adjustment of transparency in the given image. The numbers in the brackets indicate the rank of the correct answer that our model predicts. Darker colors indicate higher attention scores. More qualitative results are described in Appendix.}
\label{fig:fig4_qualitative}
\end{figure*}

\subsection{Qualitative Analysis}
To qualitatively demonstrate the advantages of our model, we visualize the attention scores of each module through examples from the VisDial v1.0 validation set in Figure \ref{fig:fig4_qualitative}. 
The attention scores of the Context Matching module, highlighted in blue, show that our model selectively focuses on contextual information as the semantic intent of the question changes. The tendency for the caption ({\em{i.e.,}} H0) to receive the highest attention score implies that the caption contains global information describing the image. In addition, the top three visual contents with high attention scores in each image lead to the potential interpretation that our model is capable of explicitly align the semantic intent ({\em{i.e.,}} highlighted in yellow) of the question and visual contents through Modality Alignment module. In more detail, the attention scores of the dialog history, highlighted in red, indicate how our model captures topic-relevant clues through previous dialog history.\\
\indent As shown in Figure \ref{fig:fig4_qualitative}(a), comparing two examples, we see that the model no longer focuses on ``6'' and ``people'' in H0 because those words are not related to the topic of the current question  ({\em{i.e.,}}``drinking''). In the example in  Figure \ref{fig:fig4_qualitative}(b), when answering Q4 (the left dialog), the model pays more attention to the question-relevant clue such as ``women'' in H0, while no longer focusing on it when answering Q6 (the question topic changes from ``tennis outfits'' to ``background''). These qualitative results show that our model successfully pays attention to visual and textual information connected to the semantic intent of the question.

\begin{figure}[t]\centering
\includegraphics[width=0.45\textwidth]{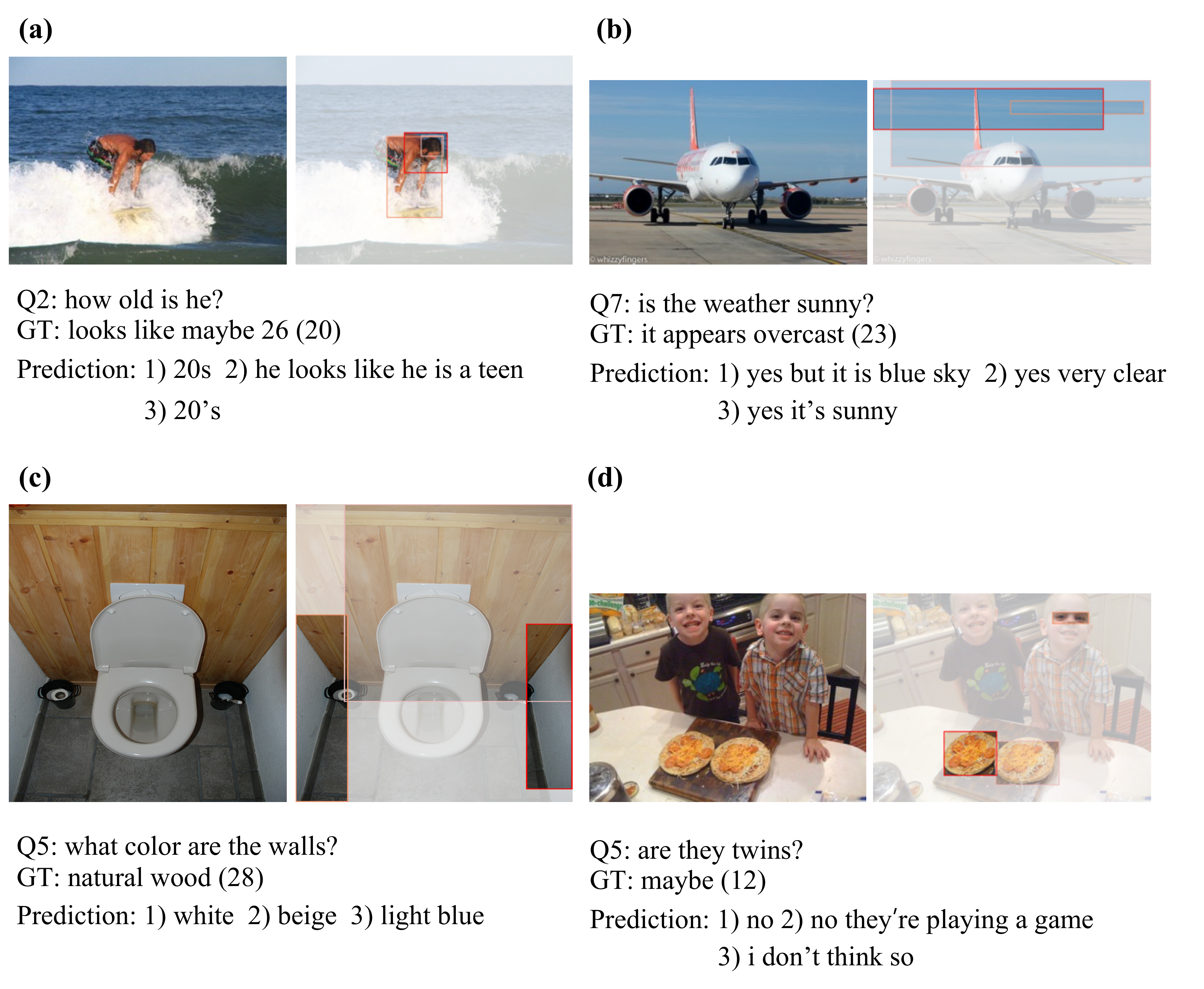}
\caption{Error analysis on the VisDial v1.0 validation set. We analyze the examples for which the model scored 0 with respect to the R@10 metric.}
\label{fig:fig5_error_analysis}
\end{figure}

\subsection{Error Analysis}
We analyzed examples in the VisDial v1.0 validation set for which our model obtained a score of 0 for the R@10 metric. The errors can be categorized into three groups: 1) Subjective judgment: our model tends to make wrong predictions for the questions about age, weather, and appearance that could involve subjective judgment, but might be acceptable (Figures \ref{fig:fig5_error_analysis}(a) and (b)). 2) Ambiguous questions: our model may focus on the wrong visual contents, for instance the left and right side walls rather than the rear wall when faced with an ambiguous question (Figure \ref{fig:fig5_error_analysis}(c)). 3) Wrong multimodal alignment: when the dialog history includes multiple entities ({\em{e.g.,}} ``boys'', ``pizzas'', and ``toppings'') that can be referenced by a single pronoun ({\em{i.e.,}} ``them''), MVAN may be confused as to which entity the pronoun refers (Figure \ref{fig:fig5_error_analysis}(d)).


\section{Conclusion}
In this paper, we introduced MVAN for the visual dialog task. MVAN can effectively determine the semantic intent of the current question and capture question-relevant information through complementary modules and sequential alignment processes. We used VisDial v1.0 to empirically evaluate our model, and as a result, our model outperforms existing state-of-the-art models. Moreover, we not only suggest plausible factors affecting a trade-off relationship of the evaluation metrics, but we enhance the interpretability of multi-level attention through detailed visualization. In future work, we aim to develop a complementary model by adding sequential information about the dialog history. Moreover, we plan to incorporate the latest pre-training methods to investigate if the performance of MVAN can be further improved. 

\bibliography{aaai2021}

\begin{figure*}[t]\centering
\includegraphics[width=0.93\textwidth]{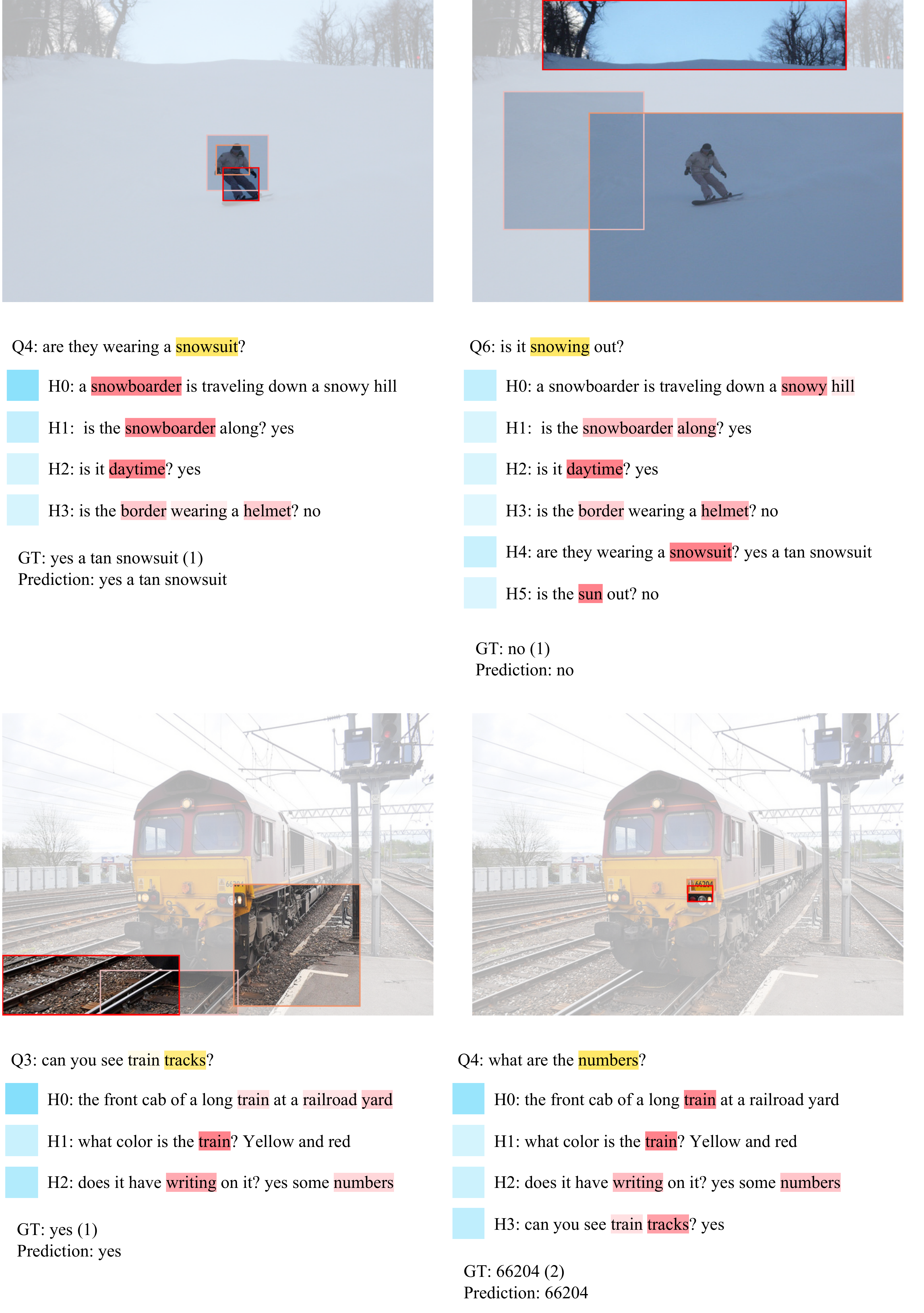}
\caption{Visualization of the reasoning process for each module in MVAN}
\label{fig:additional_v1}
\end{figure*}

\begin{figure*}[t]\centering
\includegraphics[width=0.93\textwidth]{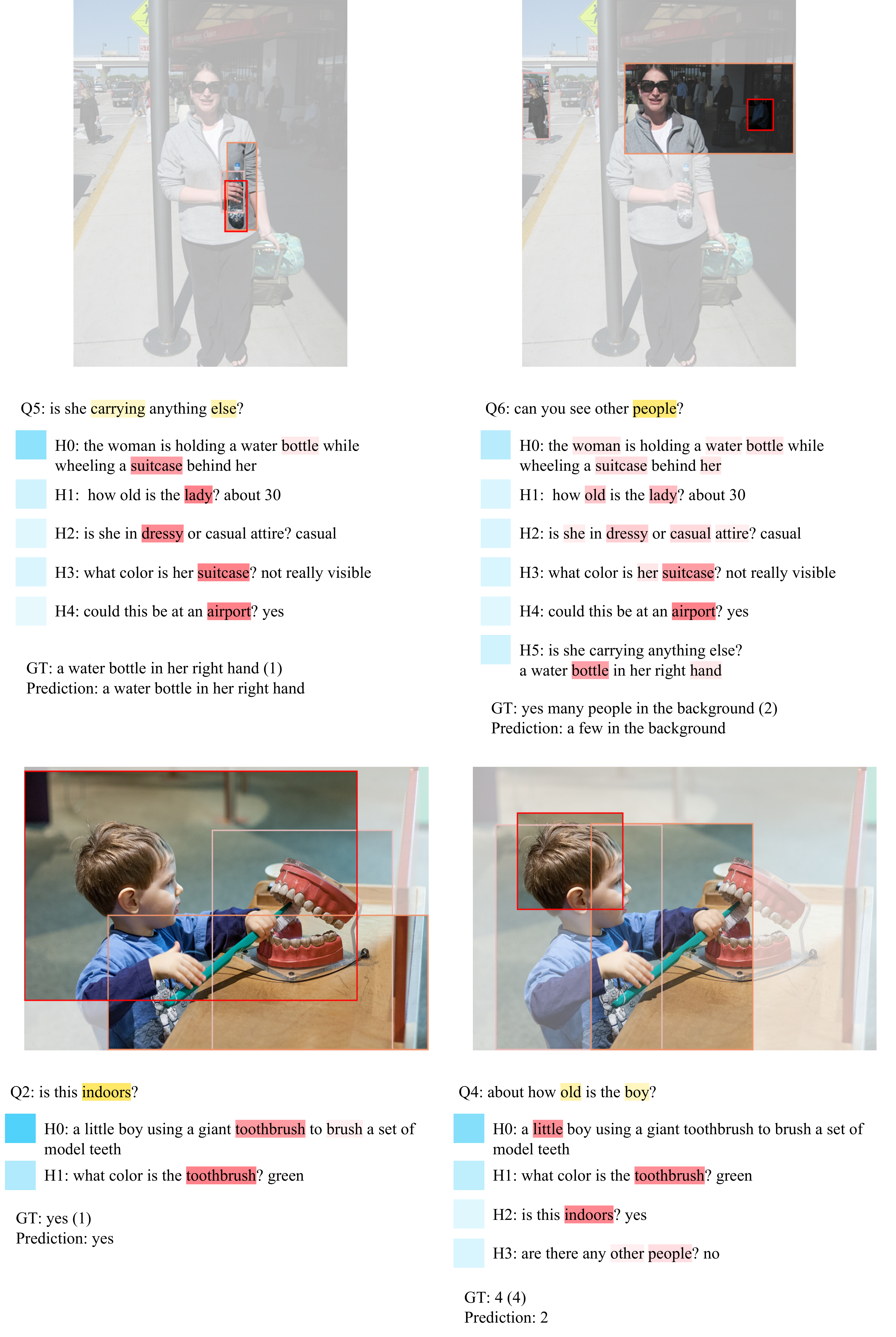}
\caption{Visualization of the reasoning process for each module in MVAN}
\label{fig:additional_v2}
\end{figure*}

\end{document}